\documentclass[11pt,a4paper]{article}
\usepackage[hyperref]{acl2019}
\usepackage{times}
\usepackage{latexsym}
\usepackage{algorithm}
\usepackage{algpseudocode}
\usepackage{amsfonts}
\usepackage[english]{babel}
\usepackage{amsthm}
\usepackage{caption}
\usepackage{subcaption}
\usepackage{amsmath}
\usepackage{tikz}
\usepackage{multirow}
\usetikzlibrary{arrows.meta,chains,decorations.pathreplacing}
\usetikzlibrary{matrix,backgrounds,positioning}

\theoremstyle{definition}
\newtheorem{exmp}{Example}[section]

\theoremstyle{definition}
\newtheorem{definition}{Definition}[section]

\usepackage{url}
\usepackage{easyReview}

\aclfinalcopy 

\title{Multi-Element Long Distance Dependencies: Using SP\emph{k} Languages to Explore the Characteristics of Long-Distance Dependencies}

\author{Abhijit Mahalunkar \\
  Applied Intelligence Research Center \\
  Technological University Dublin \\
  Dublin, Ireland \\
  \texttt{abhijit.mahalunkar@mydit.ie} \\\And
  John D. Kelleher \\
  ADAPT Research Center \\
  Technological University Dublin \\
  Dublin, Ireland \\
  \texttt{john.d.kelleher@dit.ie} \\}

\date{}

\begin{document}
\maketitle
\begin{abstract}
In order to successfully model Long Distance Dependencies (LDDs) it is necessary to understand the full-range of the characteristics of the LDDs exhibited in a target dataset. In this paper, we use Strictly \emph{k}-Piecewise languages to generate datasets with various properties. We then compute the characteristics of the LDDs in these datasets using mutual information and analyze the impact of factors such as (i) \emph{k}, (ii) length of LDDs, (iii) vocabulary size, (iv) forbidden subsequences, and (v) dataset size. This analysis reveal that the number of interacting elements in a dependency is an important characteristic of LDDs. This leads us to the challenge of modelling multi-element long-distance dependencies. Our results suggest that attention mechanisms in neural networks may aide in modeling datasets with multi-element long-distance dependencies. However, we conclude that there is a need to develop more efficient attention mechanisms to address this issue.
\end{abstract}

\section{Introduction}
Long Distance Dependencies (LDDs) describe an interaction between two (or more) elements in a sequence that are separated by an arbitrary number of positions. LDDs are related to the rate of decay of statistical dependence of two points with increasing time interval or spatial distance between them. For example, in English there is a requirement for subjects and verbs to agree, compare: ``\emph{The \textbf{dog} in that house \textbf{is} aggressive}" with ``\emph{The \textbf{dogs} in that house \textbf{are} aggressive}". This dependence can be computed using information theoretic measure i.e. \emph{Mutual Information} \cite{Cover1991,Paninski2003,Bouma2009,Lin2017}.

To date most research on LDDs has focused on the distance the dependency spans within the sequence. However, as our analysis will show the complexity of LDDs not only arises from the distance but also a number of other factors, including: (i) the number of unique symbols in a dataset, (ii) the size of the dataset, (iii) the number of interacting symbols within an LDD, and (iv) the distance between the interacting symbols. In this paper we use SPk languages to explore the complexity of LDDs. The motivation for using the SP\emph{k} language modelling task, is that the standard sequential benchmark datasets provide little to no control over the factors which directly contribute to LDD characteristics. By contrast, using SP\emph{k} languages we can generate benchmark datasets with varying degrees of LDD complexity by modifying the grammar of the SP\emph{k} language \cite{Rogers2010,Fu2011,Avcu2017}.

One aspect of LDDs that has been neglected in the research on LDDs is the complexity that arises from a multi-element dependency (i.e., dependencies that involves interactions between more than 2 elements). By controlling \emph{k} in the SP\emph{k} grammar, it is possible to generate datasets with varying degrees of multi-element dependency. This multi-element dependencies pose specific challenges to neural architectures that may require these architectures to be augmented with pointer or attention mechanisms. We explore whether attention mechanism can help with multi-element LDDs using two models, Transformer-XL \cite{dai2019} and AWD-LSTM \cite{merity2018}. Transformer-XL employs multi-head attention mechanism along with recurrence mechanism. Whereas, AWD-LSTM is a weight dropped LSTM which does not employ any attention/pointer mechanism.

The Transformer-XL and AWD-LSTM models are both \emph{language models}. A language model accepts a sequence of symbols and predicts the next symbol in the sequence. The accuracy of a language model is dependent on the capacity of the model to capture the LDDs in the data on which it is evaluated. The standard evaluation metric for language models is \emph{perplexity}. Perplexity is the measurement of the confusion or uncertainty of a language model as it predicts the next symbol in a sequence, and the lower the perplexity of a model the better the performance of the model. 


\section{Related Work: Neural Networks and Artificial Grammars}\label{sec:rel}
Formal Language Theory, primarily developed to study the computational basis of human language is now being used extensively to analyze any rule-governed system \cite{Chomsky1956,CHOMSKY1959,Fitch2012}. Formal languages have previously been used to train RNNs and investigate their inner workings. The Reber grammar \cite{REBER1967} was used to train various 1\textsuperscript{st} order RNNs \cite{6796678,118646}. The Reber grammar was also used as a benchmarking dataset for LSTM models \cite{hochreiter1997lstm}. Regular languages, studied by Tomita \cite{Tomita1982}, were used to train 2\textsuperscript{nd} order RNNs to learn grammatical structures of the strings \cite{Watrous1991,Giles1992}.

Regular languages are the simplest grammars (type-3 grammars) within the Chomsky hierarchy which are driven by regular expressions. For neural network research an interesting subclass of regular languages is the Strictly \emph{k}-Piecewise languages. Strictly \emph{k}-Piecewise languages are natural and can express some of the kinds of LDDs found in natural languages \cite{Jager2012,Heinz2010}. This presents an opportunity of using SP\emph{k} grammar to generate benchmarking datasets \cite{Avcu2017,Mahalunkar2018}. In \citet{Avcu2017}, LSTM networks were trained to recognize valid strings generated using SP\emph{2}, SP\emph{4}, SP\emph{8} grammar. LSTM could recognize valid strings generated using SP\emph{2} and SP\emph{4} grammars but struggled to recognize strings generated using SP\emph{8} grammar, exposing the performance bottleneck of LSTM networks. It has also been observed
that the performance of LSTMs on SP\emph{2} datasets degraded when the length of the LDDs in the datasets were increased, this was done by increasing the maximum length of the generated strings of SP\emph{2}  \cite{Mahalunkar2018}.

\section{Preliminaries}
\subsection{Strictly \emph{k}-Piecewise Languages (SP\emph{k})}\label{sec:spk_def}
SP\emph{k} languages form a subclass of regular languages. Subregular languages can be identified by mechanisms much less complicated than Finite-State Automata. Many aspects of human language such as local and non-local dependencies are similar to subregular languages \cite{Jager2012}. More importantly, there are certain types of long distance (non-local) dependencies in human language which allow finite-state characterization \cite{Heinz2010}. These type of LDDs can easily be characterized by SP\emph{k} languages and can be easily extended to other processes.

\begin{figure*}
\centering
\includegraphics[scale=0.4]{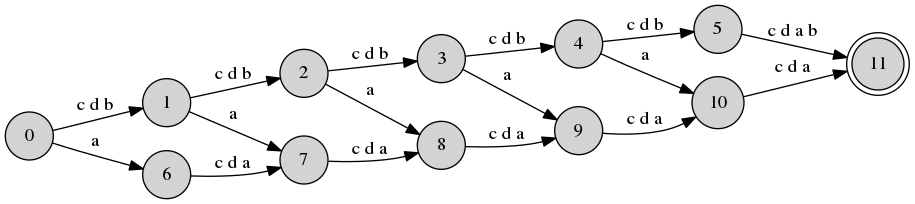}
\caption{The automaton for \emph{G\textsubscript{SP2}} where \emph{n\textsubscript{l}=6}}
\label{fig:fsa1}
\end{figure*}

A language \emph{L}, is described by a finite set of unique symbols \( \Sigma \) and \( \Sigma \)* (\emph{free monoid}) is a set of finite sequences or strings of zero or more elements from \(\Sigma \).

\theoremstyle{definition}
\begin{exmp}
Consider, \(\Sigma \) = \{\emph{\( \sigma \)\textsubscript{1}, \( \sigma \)\textsubscript{2}, \( \sigma \)\textsubscript{3}, \( \sigma \)\textsubscript{4}}\} where \emph{\( \sigma \)\textsubscript{1}, \( \sigma \)\textsubscript{2}, \( \sigma \)\textsubscript{3}, \( \sigma \)\textsubscript{4}} are the unique symbols. A \emph{free monoid} over \(\Sigma \) contains all concatenations of these unique symbols. Thus, \( \Sigma \)* = \{\emph{\( \lambda \), \( \sigma \)\textsubscript{1}, \( \sigma \)\textsubscript{1}\( \sigma \)\textsubscript{2}, \( \sigma \)\textsubscript{1}\( \sigma \)\textsubscript{3}, \( \sigma \)\textsubscript{1}\( \sigma \)\textsubscript{4}, \( \sigma \)\textsubscript{3}\( \sigma \)\textsubscript{2}, \( \sigma \)\textsubscript{3}\( \sigma \)\textsubscript{1}\( \sigma \)\textsubscript{3}, \( \sigma \)\textsubscript{2}\( \sigma \)\textsubscript{1}\( \sigma \)\textsubscript{4}\( \sigma \)\textsubscript{3}, ... }\}.
\end{exmp}

\theoremstyle{definition}
\begin{definition}
Let, \emph{u} denote a string, e.g. \emph{u}= \( \sigma \)\textsubscript{3}\( \sigma \)\textsubscript{2}. The length of a string \emph{u} is denoted by \( \vert u \vert \), and if \emph{u}= \( \sigma \)\textsubscript{3}\( \sigma \)\textsubscript{2} then \( \vert u \vert \)=2. A string with length zero is denoted by \( \lambda \).
\end{definition}

\theoremstyle{definition}
\begin{definition}\label{def1}
A string \emph{v} is a \emph{subsequence} of string \emph{w}, iff \emph{v} = \( \sigma \)\textsubscript{1}\( \sigma \)\textsubscript{2} ... \( \sigma \)\textsubscript{n} and \emph{w} \( \in \Sigma \)*\( \sigma \)\textsubscript{1}\( \Sigma \)*\( \sigma \)\textsubscript{2}\( \Sigma \)* ... \( \Sigma \)*\( \sigma \)\textsubscript{n}\( \Sigma \)*, where \( \sigma \in \Sigma\). A \emph{subsequence} of length \emph{k} is called a \emph{k-subsequence}. Let subseq\textsubscript{\emph{k}}(\emph{w}) denote the set of subsequences of \emph{w} up to length \emph{k}.
\end{definition}

\theoremstyle{definition}
\begin{exmp}
Consider, \( \Sigma \) = \{\emph{a, b, c, d}\}, \emph{w} = [\emph{acbd}], \emph{u} = [\emph{bd}], \emph{v} = [\emph{acd}] and \emph{x} = [\emph{db}]. String \emph{u} is a \emph{subsequence} of length \emph{k} = 2 or \emph{2-subsequence} of \emph{w}. String \emph{v} is a \emph{3-subsequence} of \emph{w}. However, string \emph{x} is \emph{not a subsequence} of \emph{w} as it does not contain [\emph{db}] subsequence.
\end{exmp}

SP\emph{k} languages are defined by grammar \emph{G}\textsubscript{\emph{SPk}} as a set of permissible \emph{k}-\emph{subsequences}. Here, \emph{k} indicates the number of elements in a dependency. Datasets generated to simulate 2 elements in a dependency will be generated using SP\emph{2}. This is the simplest dependency structure. There are more complex chained-dependency structures which require higher \emph{k} grammars.

\theoremstyle{definition}
\begin{exmp}\label{exp1}
Consider \emph{L}, where \( \Sigma \) = \{\emph{a, b, c, d}\}. Let \emph{G\textsubscript{SP\emph{2}}} be SP\emph{k} grammar which is comprised of permissible \emph{2-subsequences}. Thus, \emph{G\textsubscript{SP\emph{2}}} = \{\emph{aa, ac, ad, ba, bb, bc, bd, ca, cb, cc, cd, da, db, dc, dd}\}. \emph{G\textsubscript{SP\emph{2}}} grammar is employed to generate SP\emph{2} datasets.
\end{exmp}

\theoremstyle{definition}
\begin{definition}
Subsequences which are not in the grammar \emph{G} are called \emph{forbidden subsequences}\footnote{Refer section \emph{5.2. Finding the shortest forbidden subsequences} in \cite{Fu2011} for method to compute \emph{forbidden sequences} for SP\emph{k} language}.
\end{definition}

\theoremstyle{definition}
\begin{exmp}
Consider Example~\ref{exp1}, although \{\emph{ab}\} is a possible \emph{2-subsequence}, it is not part of the grammar \emph{G\textsubscript{SP\emph{2}}}. Hence, \{\emph{ab}\} is a \emph{forbidden subsequence}.
\end{exmp}

\theoremstyle{definition}
\begin{exmp}\label{exp2}
Consider strings \emph{u, v, w}: \emph{u} = [\emph{bbcbdd}], \emph{v} = [\emph{bbdbbbcbddaa}] and \emph{w} = [\emph{bbabbbcbdd}], where \( \vert \)\emph{u}\( \vert \) = 6, \( \vert \)\emph{v}\( \vert \) = 12 and \( \vert \)\emph{w}\( \vert \) = 10. Strings \emph{u} and \emph{v} are valid SP\emph{2} strings because they are composed of subsequences that are in \emph{G\textsubscript{SP\emph{2}}}. However, \emph{w} is an invalid SP\emph{2} string because \emph{w} contains \{\emph{ab}\} a subsequence which is a \emph{forbidden subsequence}. These constraints apply for any string \emph{x} where \( \vert \)\emph{x}\( \vert \in \mathbb{Z} \).
\end{exmp}

\theoremstyle{definition}
\begin{exmp}\label{LDDLength}
Let \emph{G\textsubscript{SP\emph{3}}}  = \{\emph{aaa, aab, abb, baa, bab, bba, bbb, ...}\} and \emph{forbidden subsequence} = \{\emph{aba}\} be an SP\emph{3} grammar which is comprised of permissible $3$-\emph{subsequences}. Thus, \emph{u} = [\emph{aaaaaaab}], where \( \vert \)\emph{u}\( \vert \) = 8 is a valid SP\emph{3} string and \emph{v} = [\emph{aaaaabaab}], where \( \vert \)\emph{v}\( \vert \) = 9 is an invalid SP\emph{3} string as defined by the grammar \emph{G\textsubscript{SP\emph{3}}}.
\end{exmp}

The maximum extent of LDD exhibited by a certain SP\emph{k} language is equal to the length of the strings generated which abide by the grammar. However, as per definition~\ref{def1}, the strings generated using this method will also exhibit dependencies of shorter lengths. It should be noted that the length of the LDD is not the same as \emph{k}. The length of the LDD is the maximum distance between two elements in a dependency, whereas \emph{k} specifies the number of elements in the dependency (as defined in the the SP\emph{k} grammar).

\begin{exmp}
As per Example~\ref{exp2}, \emph{v} = [\emph{bbdbbbcbddaa}], consider \emph{b} in the first position, \emph{subsequence} \{\emph{ba}\} exhibits dependency of 10 and 11. Similarly, \emph{subsequence} \{\emph{bd}\} exhibits dependency of 2, 9, 10, etc.
\end{exmp}

Figure~\ref{fig:fsa1} depicts a finite-state diagram of \emph{G\textsubscript{SP2}}, which generates strings of synthetic data. Consider a string \emph{x} from this data, \( \forall \) generated strings \emph{x} generated using grammar \emph{G\textsubscript{SP2}}: \( \vert \)\emph{x}\( \vert \) = $6$. The \emph{forbidden subsequence} for this grammar is \{\emph{ab}\}. Since \{\emph{ab}\} is a \emph{forbidden subsequence}, the state diagram has no path (from state $0$ to state $11$) because such a path would permit the generation of strings with \{\emph{ab}\} as a subsequence, \emph{e.g.} \{\emph{abcccc}\} 
Traversing the state diagram generates valid strings \emph{e.g.} \{\emph{accdda, caaaaa}\}.

Various \emph{G\textsubscript{SP\emph{k}}} could be used to define an SP\emph{k} depending on the set of \emph{forbidden subsequences} chosen. Thus, we can construct rich datasets with different properties for any SP\emph{k} language. \emph{forbidden subsequences} allow for the elimination of certain logically possible sequences while simulating a real world dataset where the probability of occurrence of that particular sequence is highly unlikely. Every SP\emph{k} grammar is defined with at least one \emph{forbidden subsequence}.

\subsection{Plotting LDD Characteristics}\label{sec:ldd_char}
Mutual information has previously been used to analyse LDDs in datasets. For example, in \citet{Ebeling2002}, mutual information was used to analyse the maximum length of the LDDs in two English literary texts, Moby Dick by H. Melville and Grimm's tales. Another example, is the work of \citet{Lin2017} who analyzed the LDD characteristics in \emph{enwik8} dataset.

Mutual information measures dependence between random variables $X$ and $Y$. These random variables have marginal distributions $p(x)$ and $p(y)$ and are jointly distributed as $p(x,y)$ \cite{Cover1991,Li1990}. Mutual information, $I(X;Y)$ is defined as;
\begin{equation}
\begin{aligned}
I(X;Y) = \sum_{x,y} p(x,y) \log \frac{p(x,y)}{p(x)p(y)}
\end{aligned}
\end{equation}
If $X$ and $Y$ are not correlated, in other words if they are independent to each other, then $p(x)p(y)=p(x,y)$ and $I(X;Y)=0$. However, if $X$ and $Y$ are fully dependent on each other, then $p(x)=p(y)=p(x,y)$ which results in the maximum value of $I(X;Y)$.

Mutual information can also be expressed using the \emph{entropy} of $X$ and $Y$ i.e. $H(X)$, $H(Y)$ and their \emph{joint entropy}, $H(X,Y)$ as given in the equations below:
\begin{equation}
\begin{aligned}
I(X;Y) = H(X) + H(Y) - H(X,Y)
\label{eq:mut-inf-h}
\end{aligned}
\end{equation}
\begin{equation}\label{shannon}
\begin{aligned}
H(X) = -\sum_{x} p(x) \log {p(x)}
\end{aligned}
\end{equation}
In our algorithm, we compute the $H(X)$ using Grassberger's corrections \cite{2003physics7138G}.
\begin{equation}
H(X) = \log N - 1/N \sum_{i=1}^{k} N_i \psi(N_i)
\label{eq:entropy-adj}
\end{equation}
where $N_i$ is the frequency of unique symbol \emph{i}, $N = \sum N_i$, $K$ is the number of unique symbols, and $\psi(N_i)$ is the \emph{logarithmic derivative of the gamma function} of $N_i$.

In order to measure dependence between any two symbols at a distance $D$ in a sequence, we design random variables $X$ and $Y$ so that $X$ holds the subsequence of the original sequence from index $0$ till $|dataset|-1-D$, and $Y$ holds the subsequence from index $D$ till $|dataset|-1$; where $D$ represents spacing between the symbols and $|dataset|$ or $LEN$ is the size of the dataset. Next we define a random variable $XY$ that contains a sequence of paired symbols one from $X$ and one from $Y$, where the symbols in a pair have the same index in $X$ and $Y$. Algorithm~\ref{ldd_algo} explains the details.


Using this information, and Equations \ref{eq:mut-inf-h} and \ref{eq:entropy-adj}, we calculate the mutual information $I(X,Y)$ at a distance $D$ in a sequence. We define the \emph{LDD characteristics} of any given sequential dataset as a function of mutual information $I(X;Y)$ over the distance $D$. Once we have calculated the mutual information within a dataset at the different distances $D$ which range between $1$ to $|dataset|$ we can then plot the LDD characteristics as a graph of distance $D$ versus mutual information at $D$. The LDD characteristics are plotted on a \emph{log-log axis}, the \emph{x-axis} defines the distance between a pair of symbols and the \emph{y-axis} marks the mutual information. 
\begin{algorithm}
\begin{algorithmic}
 \For{$D\gets 1, |dataset|$}
  \State $X \gets dataset[0:|dataset|-D]$
  \State $Y \gets dataset[D:|dataset|]$
  \State $XY \gets$ zero-matrix of size ($K^X,K^Y$)
  \For{$i\gets 0,|X|$}
    \State Increment $XY[X[i],Y[i]]$
  \EndFor
  \State Compute $N_i^X$, $N^X$, $K^X$ for $X$
  \State Compute $N_i^Y$, $N^Y$, $K^Y$ for $Y$
  \State Compute $N_i^{XY}$, $N^{XY}$, $K^{XY}$ for $XY$
  \State Compute $H(X)$, $H(Y)$ and $H(X,Y)$ Eq.~\ref{eq:entropy-adj}
  \State $I[D]\gets H(X)+H(Y)-H(X,Y)$
 \EndFor
\caption{Computing LDD Characteristics}\label{ldd_algo}
\end{algorithmic}
\end{algorithm}

\section{LDD characteristics of SP\emph{k} datasets}\label{sec:spk_ldd}
Natural datasets present little to no control over the factors that affect LDDs. This, limits our ability to understand LDDs in more detail. SP\emph{k} languages exhibit some types of LDDs occurring in natural datasets. Moreover, by modifying the SP\emph{k} grammar we can control the LDD characteristics within a dataset generated by the grammar. To understand and validate the interaction between an SP\emph{k} grammar and the characteristics of the data it generates, we used a number of datasets of SP\emph{k} grammar and analyzed the properties of these datasets. Every dataset is a collection of strings and these strings strictly follow the grammar. Hence the size of the dataset ($|dataset|$) is the sum of the size of all the strings. The datasets were generated using \emph{foma} \cite{hulden2009} and \emph{python} \cite{Avcu2017,Mahalunkar2018}\footnote{The scripts and details of these datasets are available at \url{https://github.com/silentknight/DelFol-ACL-2019}}. Below we analyze the impact of various factors on the resulting LDD characteristics.

\subsection{Impact of \emph{k}}\label{sec:k}
A dependency may arise within a dataset due to two or more interacting elements. A two element dependency is the simplest dependency structure and is analyzed by models addressing LDDs. However, multiple element dependency may not be rare and if not modeled correctly may well contribute to the errors of a model. Lack of knowledge of these dependency structures within benchmark datasets present significant limitation in comparing the performance of different models aimed at addressing LDDs. Different model architectures may be able to represent one type of LDD characteristic to a greater extent than another (e.g., distance versus \emph{k}). However, unless the experimenter is able to control the LDD characteristics present in a dataset it is not possible to disentangle which characteristics a given model struggles with based solely on the models performance on the data. SP\emph{k} grammar addresses this problem by providing control over both dependency distance and \emph{k}.

\begin{figure}
\centering
\includegraphics[scale=0.4]{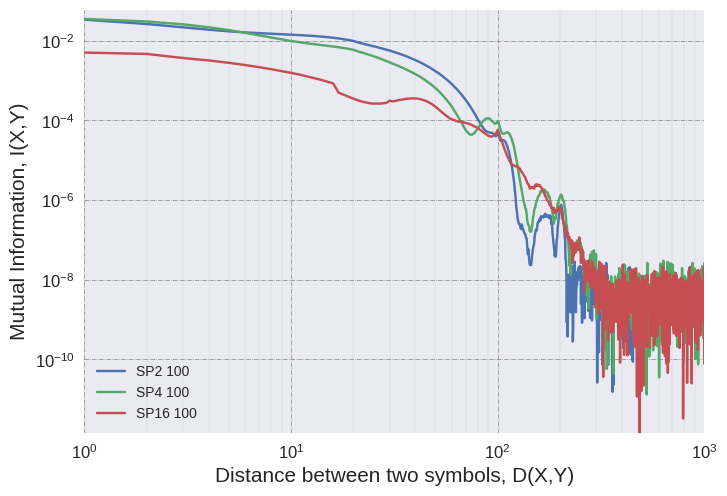}
\caption{LDD characteristics of datasets of SP\emph{2}, SP\emph{4} and SP\emph{16} grammar exhibiting LDD of length 100.}
\label{fig:spk_k}
\end{figure}

We use SP\emph{2}, SP\emph{4} and SP\emph{16} grammars to generate a set of datasets. With \emph{k}${=}\{2, 4, 16 \}$, we generate datasets with different dependency structures with interacting elements $2, 4,$ and $16$ respectively. Figure~\ref{fig:spk_k} plots the LDD characteristics of SP\emph{2}, SP\emph{4} and SP\emph{16} grammars. They contain uniform distribution of strings of string length $l$ where $60 {\leq} l {\leq} 100$. The maximum length of strings in each of these datasets is $100$. Hence, we can observe steeper mutual information decay beyond $D{>}100$. \emph{k} defines the number of correlated or dependent elements in a dependency rule. As \emph{k} increases the grammar becomes more complex and there is an overall reduction in frequency of the dependent elements in a given sequence (due to lower probability of these elements occurring in a given sequence). Hence, the mutual information is lower. This can be seen with dataset of SP\emph{16} as compared to SP\emph{2} and SP\emph{4}. It is worth noting that datasets with lower mutual information curves tend to present more difficulty during modeling \cite{Mahalunkar2018}.

\subsection{Impact of LDD length}
The distance or length between two interacting elements present significant challenge in modeling LDDs as the model is required to store the context of the interacting element persistently. The success of a model is dependent on whether it is capable of storing the required length of the contexts as dictated by the dataset.

We generated strings of maximum length $20$ $(2 {\leq} l {\leq} 20)$, $100$ $(21 {\leq} l {\leq} 100)$, $200$ $(101 {\leq} l {\leq} 200)$ and $500$ $(201 {\leq} l {\leq} 500)$ using SP\emph{2} grammar. As explained in Example~\ref{LDDLength}, by increasing the length of the generated strings, the distance between dependent elements is also increased, resulting in longer LDDs. Consequently, using this string lengths we can simulate LDD lengths of $20, 100, 200$ and $500$.

\begin{figure}
\centering
\includegraphics[scale=0.4]{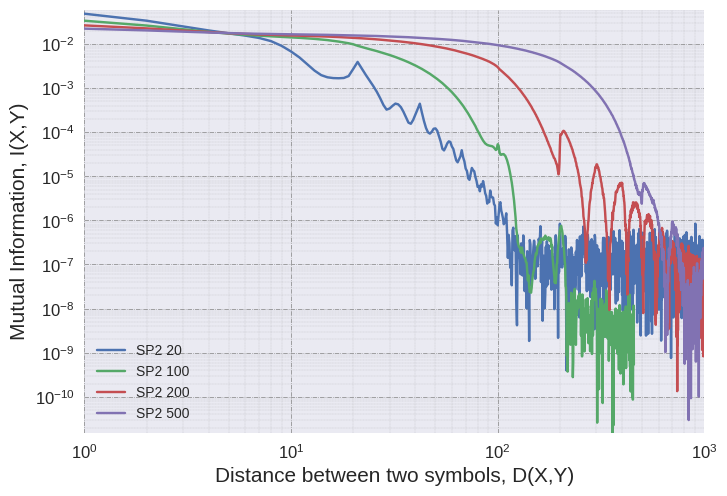}
\caption{LDD characteristics of datasets of SP\emph{2} grammar exhibiting LDDs of length 20, 100, 200 and 500.}
\label{fig:spk_len}
\end{figure}

Figure~\ref{fig:spk_len} plots LDD characteristics of SP\emph{2} languages with maximum string length of {$20, 100, 200, 500$}. The point where mutual information decay is faster, the \emph{inflection point}, lies around the same point on \emph{x-axis} as the maximum length of the LDD. This confirms that SP\emph{k} can generate datasets with varying lengths of LDDs.

\subsection{Impact of Vocabulary Size}\label{impact_v}
\begin{figure}
\centering
\includegraphics[scale=0.4]{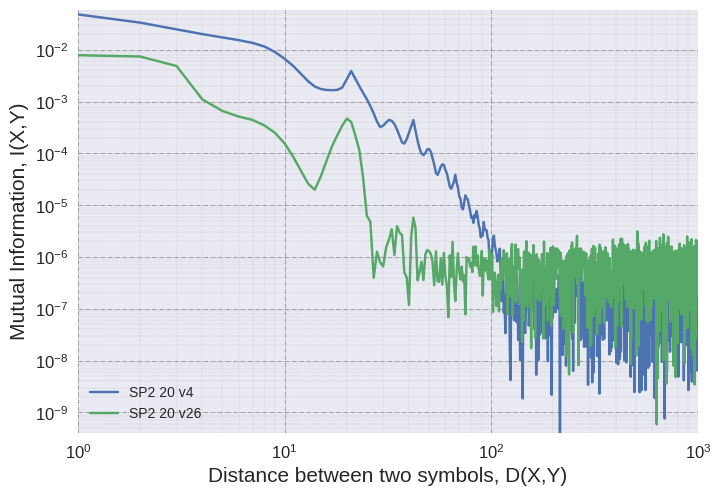}
\caption{LDD characteristics of datasets of SP\emph{2} grammar with vocabulary of 4 and 26.}
\label{fig:spk_v}
\end{figure}

We analyze the impact of vocabulary size on LDD characteristics, we generate SP\emph{2} grammars where $\Sigma_1 {=} \{ a,b,c,d \}$ ($V {=} 4$) and $\Sigma_2 {=} \{ a,b,c,d,....,x,y,z \}$ ($V {=} 26$), where $V$ is vocabulary size. The impact of vocabulary size can be seen in figure~\ref{fig:spk_v}. Both these datasets contain strings of maximum length 20. Hence the mutual information decays at 20. Both curves have identical decay indicating a similar grammar. However the overall mutual information of the dataset with $V {=} 26$ is much lower then the mutual information of the dataset with $V {=} 4$. This is because a smaller vocabulary results in an increase in the probability of the occurrence of each individual elements.

\subsection{Impact of \emph{forbidden subsequences}}
forbidden subsequences control the complexity of a given grammar. We choose two sets of forbidden subsequences for SP\emph{2} grammar, \{$ab, bc$\} and \{$ab, bc, cd, dc$\}. 

\begin{figure}
\centering
\includegraphics[scale=0.4]{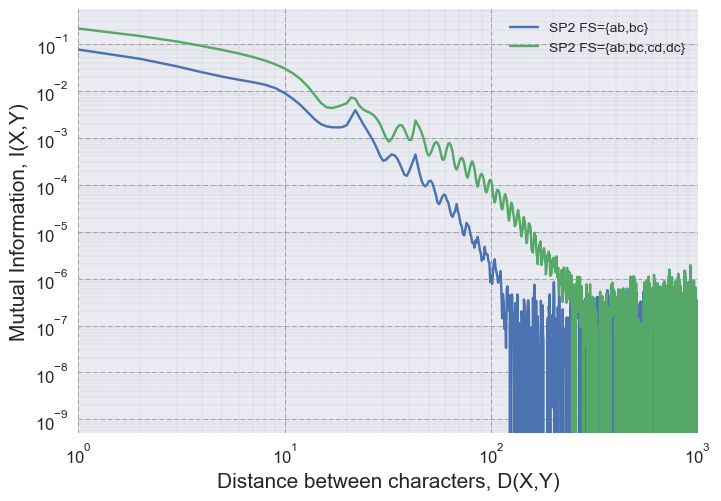}
\caption{LDD characteristics of datasets of SP\emph{2} grammar with varying forbidden subsequences.}
\label{fig:spk_f}
\end{figure}

Figure~\ref{fig:spk_f} plots the LDD characteristics of SP\emph{2} grammar with two set of \emph{forbidden subsequences} as \{$ab, bc$\} and \{$ab, bc, cd, dc$\}. It is seen that the dataset with more forbidden subsequences exhibited mutual information decay tending towards a power law decay as compared to an exponential decay by dataset with less forbidden subsequences. As explained in \citet{Lin2017}, datasets with exponential decay tend to exhibit Markovian behavior and thus are easy to model as compared to datasets with power law decay. Complex LDDs in a dataset result in power law decay. Thus, by controlling the forbidden subsequences, one can introduce more complex LDDs.

\subsection{Impact of dataset size}\label{impact_s}
Another factor to analyze is the impact of the size of the dataset ($|dataset|$) on LDDs of the same grammar. We generate two sizes of the same SP\emph{2} grammar to study the impact of the size of the data on the LDD characteristics, where one dataset is twice the size of the other.

\begin{figure}
\centering
\includegraphics[scale=0.4]{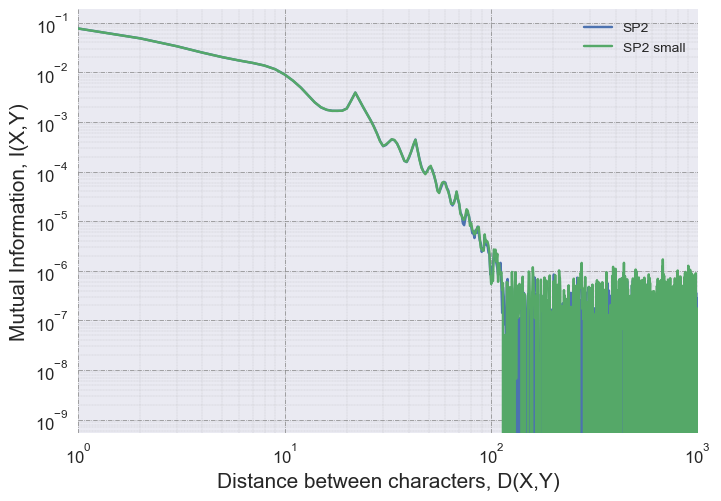}
\caption{LDD characteristics of datasets of SP\emph{2} grammar with varying size of the datasets}
\label{fig:spk_size}
\end{figure}

In figure~\ref{fig:spk_size} we can observe that LDD characteristics of datasets sampled from the same grammar are less likely to be affected by the size of the dataset.
 
\section{Multi-Element Long Distance Dependencies: Attention Mechanisms and \emph{k}}\label{sec:perplexity}
As discussed above in section~\ref{sec:k}, LDDs may arise due to multiple interacting elements, which can be referred to as multi-element long-distance dependency (ME-LDD). Current LDD research primarily focuses on developing models which are capable of retaining contextual information across long distances in sequential datasets. This approach, which focuses solely on dependency distance, may be insufficient in addressing the problems arising due to ME-LDDs. However, recent advances in attention mechanisms and memory networks may be able to represent ME-LDDs. In this section we investigate two models, Transformer-XL \cite{dai2019} (with attention mechanism) and AWD-LSTM (ASGD Weight-Dropped LSTM) (without attention mechanism) \cite{merity2018} so as to analyze how attention mechanisms help in modeling datasets with ME-LDDs.

The Transformer-XL model augments vanilla Transformer models by introducing a recurrence mechanism to the Transformer architecture. This recurrence effectively encodes an arbitrarily long context into a fixed size representation over constrained memory and computation. A vanilla Transformer is made up of Multi-Head Attention and Feed Forward layers which aides in adding positional information to the embedded representation. The other model we tested, the AWD-LSTM, uses a weight-drop mechanism so as to aid in regularization of the LSTM network. Hence this model does not explicitly uses attention mechanism.

We trained both these models with variants of SP\emph{k} grammar where \emph{k}${=}\{2,4,8,16\}$ for vocabulary size $V{=}4$ and \emph{k}${=}\{2,4,6,8\}$ for vocabulary size $V{=}26$. Hence, there are in total $8$ datasets. Every dataset contains uniform distribution of strings with string length $l$ where $60 \leq l \leq{100}$. The string length ordering is not maintained so as to not bias the models. In-order to train the language models, every dataset is split into training/validation/test sets. All the training sets for vocabulary size $V{=}4$ contain ${\approx} 195000$ number of strings and the size of the training set is ${\approx}24MB$. Similarly, all the training sets for vocabulary size $V{=}26$ contain ${\approx} 222000$ number of strings with size of ${\approx}40MB$. All the test and validation sets contain ${\approx} 16000$ strings with size of $2MB$\footnote{The datasets and scripts used for training the models can be found at \url{https://github.com/silentknight/DelFol-ACL-2019}}. The hyperparameters for both the models were reused from the \emph{Penn Treebank character} model \cite{dai2019,merity2018}.

\begin{table}
\centering
\begin{tabular}{|c||c|c|c|c|}
\hline
\multirow{2}{*}{Models} & \multicolumn{4}{c|}{Test Perplexity in \emph{bpc}} \\ \cline{2-5}
    & SP$2$ & SP$4$ & SP$8$ & SP$16$ \\ \hline \hline
$1$   & $1.6855$ & $1.8038$ & $1.9611$ & $2.0759$ \\ \hline
$2$   & $1.413$ & $1.486$ & $1.658$ & $1.708$ \\ \hline
\end{tabular}
\caption{Perplexity score of $1$: Transformer-XL and $2$: AWD-LSTM models of SP$2$, SP$4$, SP$8$ and SP$16$ datasets with vocabulary size $V{=}4$}
\label{tab:perplexity_score_v4}
\end{table}

\begin{table}
\centering
\begin{tabular}{|c||c|c|c|c|}
\hline
\multirow{2}{*}{Models} & \multicolumn{4}{c|}{Test Perplexity in \emph{bpc}} \\ \cline{2-5}
     & SP$2$ & SP$4$ & SP$6$ & SP$8$ \\ \hline \hline
$1$  & $4.6846$ & $4.7320$ & $4.7384$ & $4.7385$ \\ \hline
$2$  & $4.525$ & $4.635$ & $4.707$ & $4.708$ \\ \hline
\end{tabular}
\caption{Perplexity score of $1$: Transformer-XL and $2$: AWD-LSTM models of SP$2$, SP$4$, SP$6$ and SP$8$ datasets with vocabulary size $V{=}26$}
\label{tab:perplexity_score_v26}
\end{table}

\begin{figure}
\centering
\includegraphics[scale=0.4]{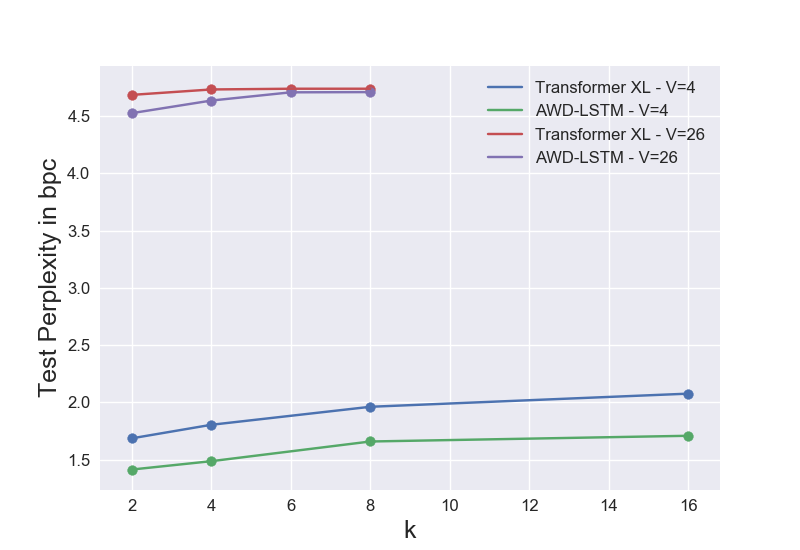}
\caption{Test perplexity score of Transformer-XL and AWD-LSTM models of SP$2$, SP$4$, SP$8$ and SP$16$ datasets.}
\label{fig:perplexity}
\end{figure}

Table~\ref{tab:perplexity_score_v4} lists the test perplexity scores of both the models in \emph{bits per character} for datasets with vocabulary size $V{=}4$ and table~\ref{tab:perplexity_score_v26} lists the test perplexity scores of both the models for datasets with vocabulary size $V{=}26$. We plot the test perplexity scores of all the datasets as function of $k$ in figure~\ref{fig:perplexity}. Here we observe two distinct sets of perplexity growth curves. It can be seen that, as the size of the vocabulary changes, the test perplexity score across all the models also changes relatively. This confirms that the vocabulary size of the dataset does impact the perplexity score of the models. This is attributed to the lower mutual information in the LDD characteristics as explained in section~\ref{impact_v}.

Switching our focus to the growth in perplexity as \emph{k} increases, we tried to understand how the presence of attention mechanism impacts the ability of the neural architectures to model the ME-LDDs as \emph{k} increases. For datasets with $V{=}26$, the impact of attention mechanism in Transformer-XL is apparent. The test perplexity score remains almost unchanged which we attribute to the presence of an attention mechanism. However, AWD-LSTM model struggles to maintain test perplexity scores as \emph{k} increases, for datasets with $V{=}26$ (higher vocabulary size).

For datasets with $V{=}4$, it can be seen that the test perplexity of Transformer-XL model as a function of \emph{k} scales exponentially. The presence of exponential relation indicates that the attention mechanism of the Transformer helps the model as \emph{k} increases in the SP\emph{k} grammar: the exponential relationship indicates that the growth in model's perplexity appears to saturate as \emph{k} increases. For AWD-LSTM model, the test perplexity with \emph{k} also increases at similar rate as that of Transformer-XL. This could be attributed to smaller vocabulary size, which leads to less complex dependency structure even at higher value of $k$. Such datasets could be easily modeled using non-attention mechanisms as seen in the figure~\ref{fig:perplexity}.

Comparing the test perplexity scores in figure~\ref{fig:perplexity} for both of these models, it can be observed that the overall test perplexity score of Transformer-XL model for across all the datasets as compared with AWD-LSTM model is higher. This could be attributed to over-parameterization of the Transformer-XL model. Transformer-XL uses ${\approx}44$ million parameters as compared to ${\approx}18$ million used by AWD-LSTM model to model these sub-regular languages. It should also be noted that the height of the LDD characteristics of all the datasets is significantly lower than natural datasets \cite{Mahalunkar2018b}. Consequently the mutual information is very small at a given distance $D$. This leads to higher perplexity in language models modeling them.

We also observed no apparent impact on the value of test perplexity scores of all the models on training sets with twice the number of strings. This can be substantiated by observing the LDD characteristics in section~\ref{impact_s}. Limitation of \emph{foma} tool to generate datasets with various properties prevented us from exploring more complex datasets. \emph{E.g.} we were unable to generate SP$16$ dataset with vocabulary size of $V{=}26$ due to \emph{stack full} error.

\section{Discussion}\label{sec:disc}
The LDD characteristics of a dataset is indicative of the presence of a certain type of grammar in the dataset. Our experiments reveal that even though a specific grammar does induce similar LDD characteristics, there are subtle variations. These variations depend on a number of factors such as size of the vocabulary, length of contextual relations, dependency structure (for e.g. ``\emph{k}" and ``\emph{forbidden subsequences}"). This analysis improves our understanding of the complex nature of the LDDs. This analysis can be extended to natural datasets in an effort to better understand the datasets. Thus, if a sequential model such as recurrent neural architecture intends to model a dataset, knowing these factors would greatly benefit in selecting the best hyper-parameters of the sequential model. By training Transformer-XL and AWD-LSTM model with datasets possessing various properties, it was possible to observe the impact of various properties on the perplexity score. Also, the impact of multi-head attention mechanism on the vocabulary size is quite evident. Our results suggest that the Transformer-XL performs much better with increase in vocabulary size. It is also evidenced by its SoTA results on \emph{WikiText103} dataset ($V{=}267735$) \cite{dai2019}.

\section{Conclusion}
The majority of neural network research on sequential models focuses solely on modelling dependencies across long distances.  However, the dependencies that occur in sequential data can also be multi-element. Furthermore, the vocabulary size, and the forbidden subsequences within a grammar also contribute to the difficulty of modelling the dependencies within a dataset.

In natural datasets all of these factors interact, and can confound the analysis for model performance. However, using SP\emph{k} languages it is possible to synthesize sequential datasets and control the type of dependencies exhibited in these datasets. Using a mutual information based analysis of SP\emph{k} synthesized datasets we examined how the different language characteristics (vocabulary size, forbidden subsequences) and dependency characteristics (length, \emph{k}) are reflected in the datasets generated by SP\emph{k} grammars. Furthermore, our results suggest that attention mechanisms in neural networks may aide in modeling datasets with multi-element long-distance dependencies. Although we encourage developing more efficient models.

The potential impact of this work for neural networks research include: an appreciation of the multifaceted nature of LDDs; a procedure for measuring LDD characteristics within a dataset; an understanding of how different hyper-parameters setting on an SP\emph{k} based dataset synthesis process (string length, \emph{k}, forbidden subsequences, vocabulary size) affect the mutual information profile of the resulting dataset.

\section*{Acknowledgment}
This research was partly supported by the ADAPT Research Centre, funded under the SFI Research Centres Programme (Grant 13/RC/2106) and is co-funded under the European Regional Development Funds. The research was also supported by an IBM Shared University Research Award. We gratefully acknowledge the support of NVIDIA Corporation with the donation of the Titan Xp GPU under NVIDIA GPU Grant used for this research.
\section*{}

\bibliography{acl2019}
\bibliographystyle{acl_natbib}
\end{document}